\documentclass[letterpaper, 10 pt, conference]{ieeeconf}
\IEEEoverridecommandlockouts
\usepackage{lscape}
\usepackage[utf8]{inputenc}
\usepackage{graphicx}
\usepackage{xcolor}
\usepackage[export]{adjustbox} 
\usepackage{amssymb}
\usepackage{algorithm}
\usepackage{algpseudocode}
\usepackage{mathtools}
\usepackage[export]{adjustbox}
\usepackage{afterpage}

\usepackage{geometry}
\usepackage[colorinlistoftodos,prependcaption,textsize=footnotesize]{todonotes}

\definecolor{reachColor}{rgb}{0.71,0.509,0.439}
\definecolor{transColor}{rgb}{0.647,0.858,0.647}
\definecolor{retreatColor}{rgb}{0.647,0.647,0.858}

\newcommand\reach[1]{\textcolor{reachColor}{\textbf{#1}}}
\newcommand\transport[1]{\textcolor{transColor}{\textbf{#1}}}
\newcommand\retreat[1]{\textcolor{retreatColor}{\textbf{#1}}}

\title{\large \bf
Intuitive \& Efficient Human-robot Collaboration via Real-time Approximate Bayesian Inference}

\author{\small Javier Felip \and \small David Gonzalez-Aguirre \and \small Lama Nachman
    \thanks{Authors are with Intel Labs, Intelligent Systems Research Lab. \{javier.felip.leon, david.i.gonzalez.aguirre, lama.nachman\}@intel.com}
}

\begin{document}

\maketitle
\thispagestyle{empty}
\pagestyle{empty}

\begin{abstract}
The combination of collaborative robots and end-to-end AI, promises flexible automation of human tasks in factories and warehouses. However, such promise seems a few breakthroughs away. In the meantime, humans and cobots will collaborate helping each other. For these collaborations to be effective and safe, robots need to model, predict and exploit human's intents for responsive decision making processes.

Approximate Bayesian Computation (ABC) is an analysis-by-synthesis approach to perform probabilistic predictions upon uncertain quantities. ABC includes priors conveniently, leverages sampling algorithms for inference and is flexible to benefit from complex models, e.g. via simulators. However, ABC is known to be computationally too intensive to run at interactive frame rates required for effective human-robot collaboration tasks.

In this paper, we formulate human reaching intent prediction as an ABC problem and describe two key performance innovations which allow computations at interactive rates. Our real-world experiments with a collaborative robot set-up, demonstrate the viability of our proposed approach. Experimental evaluations convey the advantages and value of human intent prediction for packing cooperative tasks. Qualitative results show how anticipating human's reaching intent improves human-robot collaboration without compromising safety. Quantitative task fluency metrics confirm the qualitative claims.
\end{abstract}

\section{Introduction}
Multiple user studies \cite{Palinko2016, Hoffman2019, Huang2016} show that humans continuously predict co-worker's intent when performing collaborative tasks. Such intent prediction is key for effective human-human collaboration tasks \cite{Butepage2017}. In this paper, we refer to human reaching intent prediction as the ability to project human reaching motion into future spatio-temporal regions solely based on partial noisy observations.

Human behaviours are difficult to predict and robot perception is approximate and prone to noise. Probabilistic approaches are equipped with a mathematical framework that takes uncertainty into consideration. For these reasons, probabilistic methods are very popular among the robotics community \cite{Thrun2005}. 

ABC provides a principled analysis-by-synthesis approach, capable of sampling from distributions whose likelihood is unknown (or intractable) but a good model for generating synthetic data is available \cite{Rubin1984, Pritchard1999}. Samples can later be used to perform inference with uncertainty quantification. 

ABC flexibly incorporates prior distributions exploiting observed environmental states, task descriptions, or other a priori information sources. The generative model is also a flexible design choice, for example physics simulators, render engines, Generative Adversarial Networks, etc. A similarity metric is the last big design decision. ABC rejects generated data that is not close enough to observations. The basics of ABC are introduced in Section~\ref{sec:ABC} and its application to human intent estimation is described in Section~\ref{sec:abc_intent}.

Due to its high computational cost, applying ABC to problems that require interactive frame-rates is still not a common approach. Besides, the design and application of ABC approaches to robotics problems is not straightforward, some design choices require expertise. In this paper, we map all components of ABC for human reaching intent prediction. We also show how to accelerate ABC (5 orders of magnitude) to run at interactive frame rates on desktop processors. We believe this example can be helpful for the application of ABC to many other robotics problems.

Our experimental scenario focuses on a turn-based human-robot collaboration pick and place task. A human and a robot collaborate to pick objects from a common workspace and place them into a shared box. We demonstrate that reaching intent prediction significantly improves task fluency leading to productive and safe process enhancement.

The contributions of this paper are threefold:
\begin{itemize}
    \item Formulation and implementation of a human reaching intent prediction based on ABC. We show how to design priors from task, object perception and human states. We introduce our novel generative approach for the task based on physics simulation. We demonstrate how to use the inferred intent to improve task fluency for human-robot collaborative packing.
    \item Enable ABC to run at unprecedented interactive rates. Thanks to a novel combination of neural surrogates and dynamic programming.
    \item Provide experimental evidence supporting our claims about human intent prediction improving fluency and performance in collaborative human-robot tasks.
\end{itemize} 

\section{Related work}
The foundations of ABC were set by Rubin in 1984 \cite{Rubin1984} and first used in practice by Pritchard \cite{Pritchard1999} in 1999. In the literature, they are also known as likelihood-free methods and are closely related to analysis by synthesis approaches. Since their invention, many different advances and optimizations have been proposed and demonstrated. Depending on the task at hand, it might be favorable to use a different variation of ABC. Variations are obtained by modifying the modular components of ABC (See details in Section~\ref{sec:ABC}). There are a number of libraries that are focused on helping developers apply ABC to their inference problems \cite{Wegmann2010, dutta2021abcpy, lintusaari2018elfi}. However, even with such tools, ABC is known to be computationally expensive. Hence, ABC is up to now rarely used for real-time inference problems. The techniques introduced in this paper alleviate the computational burden and can be applied to existing ABC libraries.

Intent prediction is a key capability to improve human-robot collaboration \cite{Butepage2017}. Due to the complexity associated with automatic task segmentation and time registration, most intent prediction papers do not evaluate their methods with fluency or performance metrics which makes them difficult to compare. In this paper we use key metrics identified by Hoffman \cite{Hoffman2019}. Our experimental procedure automatically segments human task execution in phases. This enables to soundly compute performance and fluency metrics.

Perez-D'Arpino and Shah used a data-driven method able to classify the reaching intent of a human on a 2x2 matrix of objects \cite{Perez_Shah2015}. Huang and Mutlu proposed a method based on detecting patterns in gaze tracking to anticipate the object being requested\cite{Huang2016}. Deep learning approaches with Recurrent Neural Networks have also been proposed \cite{gao_2021}. 

On the model-based methods, Zanchettin and Rocco \cite{zanchettin_2017} proposed an analysis-by-synthesis approach with a generative model based on minimum jerk trajectories and a handcrafted likelihood. Luo et. al. \cite{luo_2018} proposed an on-line data-driven approach that learns different motion models on the fly.

Our method is hybrid with data-driven and model-based capabilities depending on the nature of the generative model used. It can make inference over more targets, is probabilistic, application flexible, able to incorporate priors from multiple sources, and is computationally more efficient.



\section{Approximate Bayesian Computation}
\label{sec:ABC}
ABC is a principled method to infer quantities of interest with unknown posterior distributions, i.e. it is not possible to sample or its likelihood is intractable\footnote{There are multiple reasons for distributions to be intractable. Unknown normalization constant, require to compute intractable integrals, there are no algorithms to generate samples from the distribution, no closed form for the likelihood, etc.}. It requires three components: First, \emph{observations} or measurements that provide data that is related with the quantities of interest. Second, a \emph{generative model} that produces data in the observation space given values from inference target space. Third, a \emph{similarity function} which provides a proximity measure between two points in the observation space. Additionally, a \emph{sampling algorithm} is needed as part of the ABC procedure.

In essence, ABC is a synthesis-based method able to obtain samples from an unknown posterior distribution. It samples from a \emph{prior} distribution and feeds the sampled value to a \emph{generative model} which produces synthetic data. By rejecting the samples that generated synthetic data far from \emph{observations}, ABC guarantees that the retained samples are exact samples from the posterior \cite{Rubin1984}. The vanilla ABC-reject algorithm \cite{Pritchard1999} is detailed in Algorithm~\ref{alg:abc}, where the inputs $\pi(z)$ are the prior distribution, $L$ is the similarity metric, $\epsilon$ is the sample tolerance, $x$ is the current observation and $n$ is the number of desired samples.


\begin{algorithm}[t]
	\caption{ABC reject}
	\label{alg:abc}
	\begin{algorithmic}[1]
        \State {\bfseries ABC\_Reject}($\pi, L, \epsilon, x, n$)
        \State $N \leftarrow \emptyset$
		\While{$|N| < n$}
		\State $\hat{z} \sim \pi(z)$ \hfill\makebox[0.55\linewidth][l]{$\triangleright$ Sample from the prior.}
		\State $\hat{x} \sim g_{\theta}(\hat{z})$ \hfill\makebox[0.55\linewidth][l]{$\triangleright$ Generate an observation.}
		\If{$L(x, \hat{x}) < \epsilon$}
		\State $N \leftarrow N \cup \{\hat{z}\}$ \hfill\makebox[0.55\linewidth][l]{$\triangleright$ Keep $\hat{z}$. Else $\hat{z}$ is rejected.}
		\EndIf
		\EndWhile \hfill\makebox[0.55\linewidth][l]{$\triangleright$ Stop when enough samples.}
		\State {\bfseries return} $N$
	\end{algorithmic}
\end{algorithm}

 \subsection{ABC for human reaching intent prediction}
 \label{sec:abc_intent}

For predicting human intent, we define our quantity of interest $z\in \mathbb{R}^3$ as the target position on the table towards which a human is reaching. The trajectory that the human hand describes is used as the partial observation $x: t \in [0,3]  \subset \mathbb{R} \mapsto z\in \mathbb{R}^3 $ which depends on the human reaching intent. Therefore we assume that the observed hand trajectory is conditioned by the intent, having a likelihood $p(x|z)$. However, the likelihood function for human manipulation intent is unknown. Here is where ABC becomes handy, we can design a model that given a reaching intent, it can generate plausible hand trajectories $g_\theta(z) \to \hat{x}$. By observing the hand trajectory $x$, we can compare the generated data to direct observations using a similarity metric $L(x, \hat{x}) \rightarrow \mathbb{R}^+$. ABC states that if we keep the intent values $z$ that generated data $\hat{x}$ close enough to observations $L(x, \hat{x}) < \epsilon$ (up to an epsilon) those intent values are actually drawn directly from the posterior.

\subsection{Human reaching generative model}
\label{sec:gen_model}

\begin{figure}
    \centering
    \includegraphics[width=0.32\columnwidth]{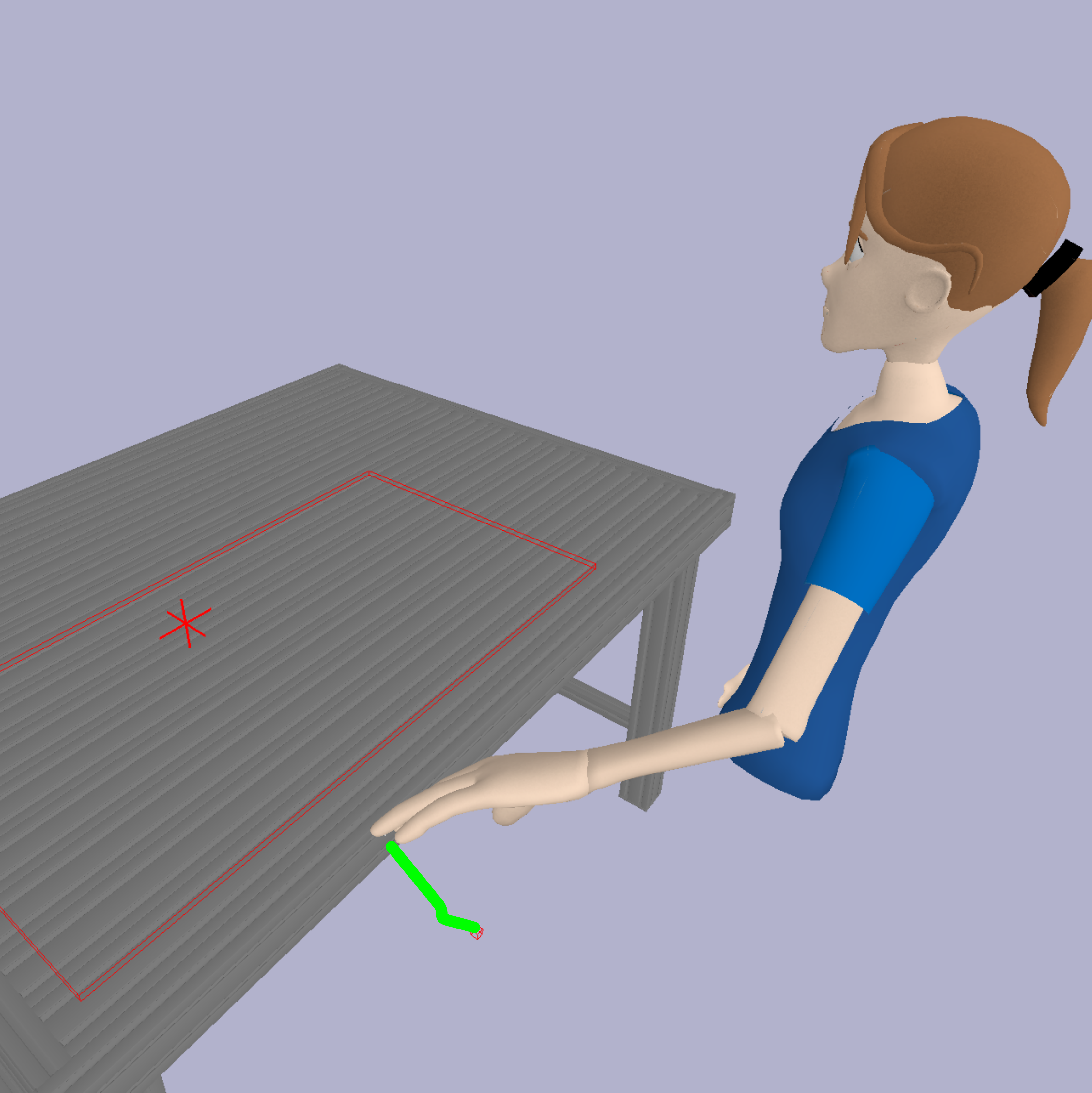}
    \includegraphics[width=0.32\columnwidth]{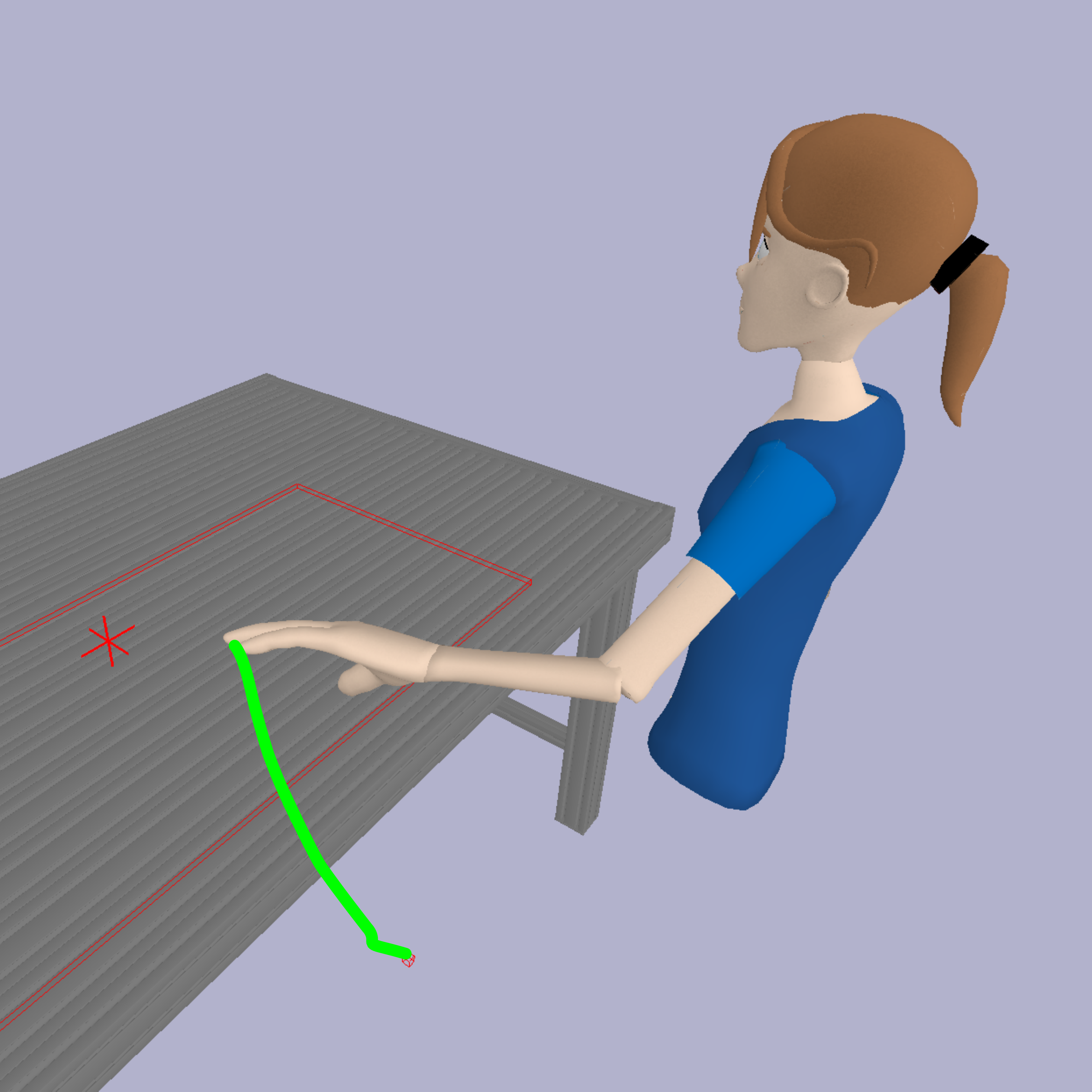}
    \includegraphics[width=0.32\columnwidth]{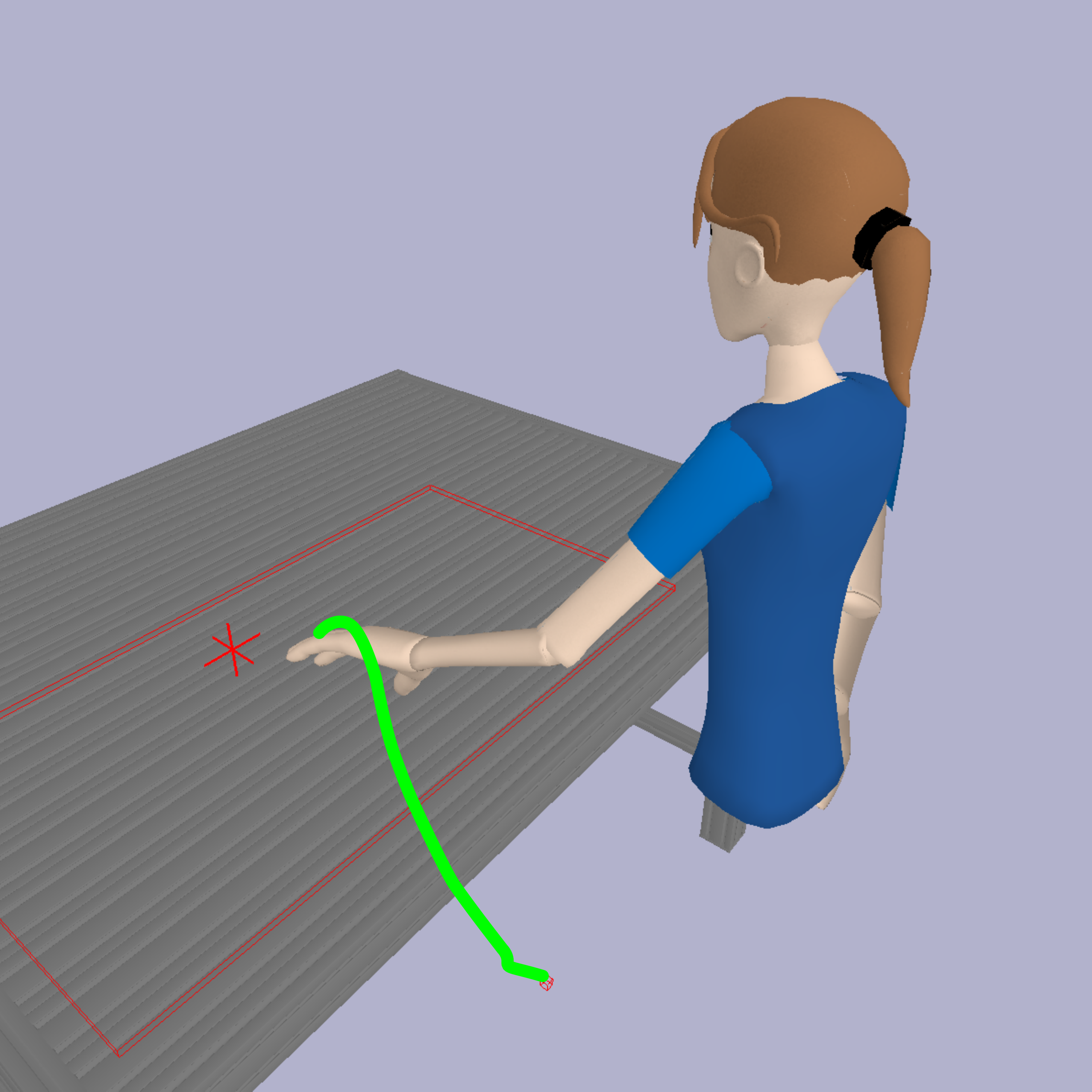}
    \caption{Human reaching trajectory generative model. From left to right, example sequence of a generated trajectory (in green) from the initial hand position to the target reaching position (red 3D crosshair). Trajectories are generated using $K_p=6, K_i=0.01, K_d=0.1, K_{rep}=8.5$.}
    \label{fig:gen_model}
\end{figure}

The generative model consists of a kinematic model of the task environment, and a 9 Degrees-of-Freedom (DoF) model of a human torso and left arm whose joint limits and dimensions are set to average US adult values\footnote{Source: National Center for Health Statistics. Anthropometric Reference Data for Children and Adults: United States, 2011-2014}. 
The kinematic model is loaded into a physics engine \cite{bullet} where trajectories are generated by simulating 3 seconds of hand trajectory at 30Hz. Resulting in trajectories of 90 tridimensional points, see generating sequence in Fig~\ref{fig:gen_model}. 

The arm motion is obtained by a Jacobian pseudo-inverse $J^{\dagger}$ that transforms task space velocities $\dot{x}$ into joint space velocities $\dot{\theta}$ (see Eq.~\ref{eq:joint_space_controller}). $\dot{x}$ is obtained by means of a PID controller (See Eq.~\ref{eq:pos_error} and \ref{eq:task_space_controller}) that attracts the hand position $x$ to the target location $x_{des}$ and a Potential Fields controller \cite{khatib86} that pushes the hand away from obstacles $x_{obj}$. The controller gains ($K_p, K_i, K_d, K_{rep}$) are heuristically tuned to provide smooth motion and goal convergence. An example generated trajectory with the tuned parameters is depicted in Figure~\ref{fig:gen_model}. The redundant degrees of freedom are used to maintain a human-like posture while the trajectory is being executed. To do so, a second joint position objective $\theta_{sec}$ is projected into the controller using the nullspace of the Jacobian pseudo-inverse $Null(J^{\dagger})$ \cite{LynchPark2017}, namely
\begin{equation}
    e(t) = x_{des}(t)-x(t)
	\label{eq:pos_error}
\end{equation}
\begin{equation}
    \dot{x}(t) = K_p e(t)+ K_i \int_0^t e(t) dt + K_d \frac{de(t)}{dt} + K_{rep}(x(t)-x_{obj}) 
	\label{eq:task_space_controller}
\end{equation}
\begin{equation}
    \dot{\theta}(t) = J^{\dagger}(t) \dot{x}(t) + \text{Null}(J^{\dagger}(t)) (\theta_{sec} - \theta(t))
	\label{eq:joint_space_controller}
\end{equation}

\subsection{Observation similarity metrics}
ABC needs a similarity measure to determine how close generated synthetic data is to the observed data points. For reaching intent prediction, observations consist of the partial trajectory of the hand while reaching for an object. The similarity is computed only to the partial generated trajectory.

Curve similarity metrics, like Hausdorff or Fr{\'e}chet distances, are good candidates to compare partially generated and observed trajectories. However, their computational overhead is a problem when real-time inference is needed. For this reason, Mean Squared Error (MSE) provides dependable results while exposing higher resiliency to perception outliers with superior computational performance.

When comparing two partially observed trajectories, the time window $w$ for the comparison needs to be considered. It determines the number of points used for the metric computation. Short time windows focus on the last observed points, can adapt faster, but are more sensitive to noise or small variations. Longer windows can capture features from further in the past and are more resilient to variations but adapt slower to abrupt changes. This results in the \emph{windowed MSE} similarity:
\begin{equation}
L(x, \hat{x}) \coloneqq || x_{t:t+w} - \hat{x}_{t:t+w} ||_2^2.
\end{equation}

Finally, the similarity threshold $\epsilon$ is a quality  (computation vs approximation) trade-off. Small values will increase the amount of samples from the prior that are rejected. High rejection rates force more sampling iterations to obtain the same amount of accepted samples. Hence, it results in improved approximation quality at an increased computational cost. Therefore the design of good prior distribution, that is as close as possible to the posterior is key for sampling efficiency.

\subsection{Designing informative prior distributions}
The focus of this section is twofold. First, demonstrate the flexibility of ABC to incorporate priors that leverage different information sources such as task structure, domain knowledge and common sense. Second, exemplify how to derive informative priors for the reaching intent inference case. Informative priors, when close to the target posterior increase the sample acceptance rate and boost the approach efficiency. There's only one requirement, a process to sample from them must exist.

\begin{figure*}
    \centering
    \includegraphics[width=0.475\columnwidth]{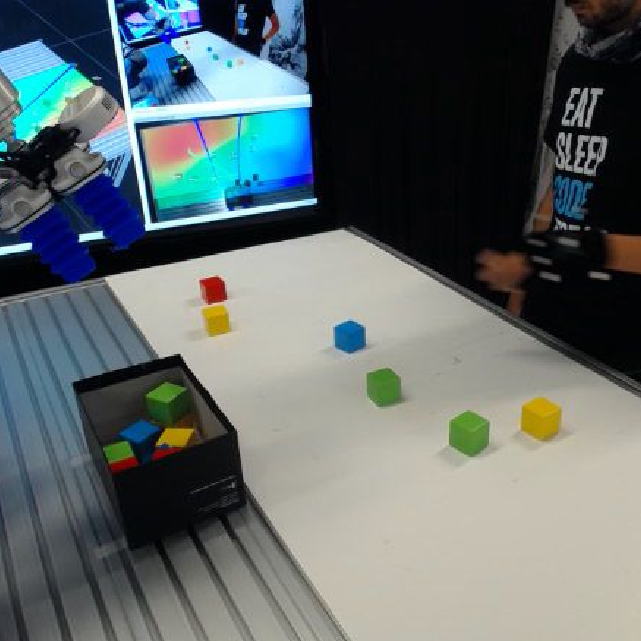}
    \includegraphics[width=0.475\columnwidth]{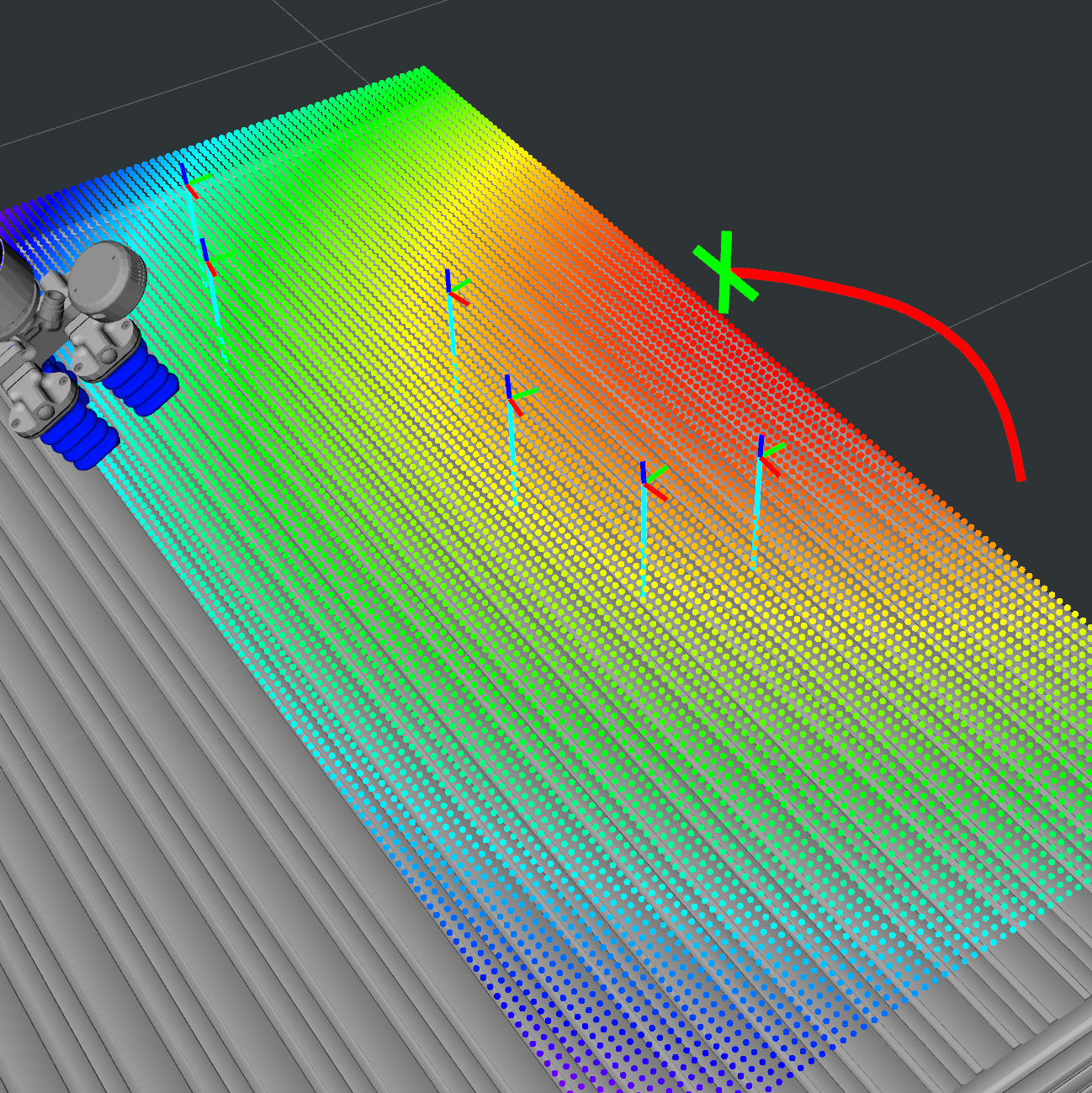}
    \includegraphics[width=0.475\columnwidth]{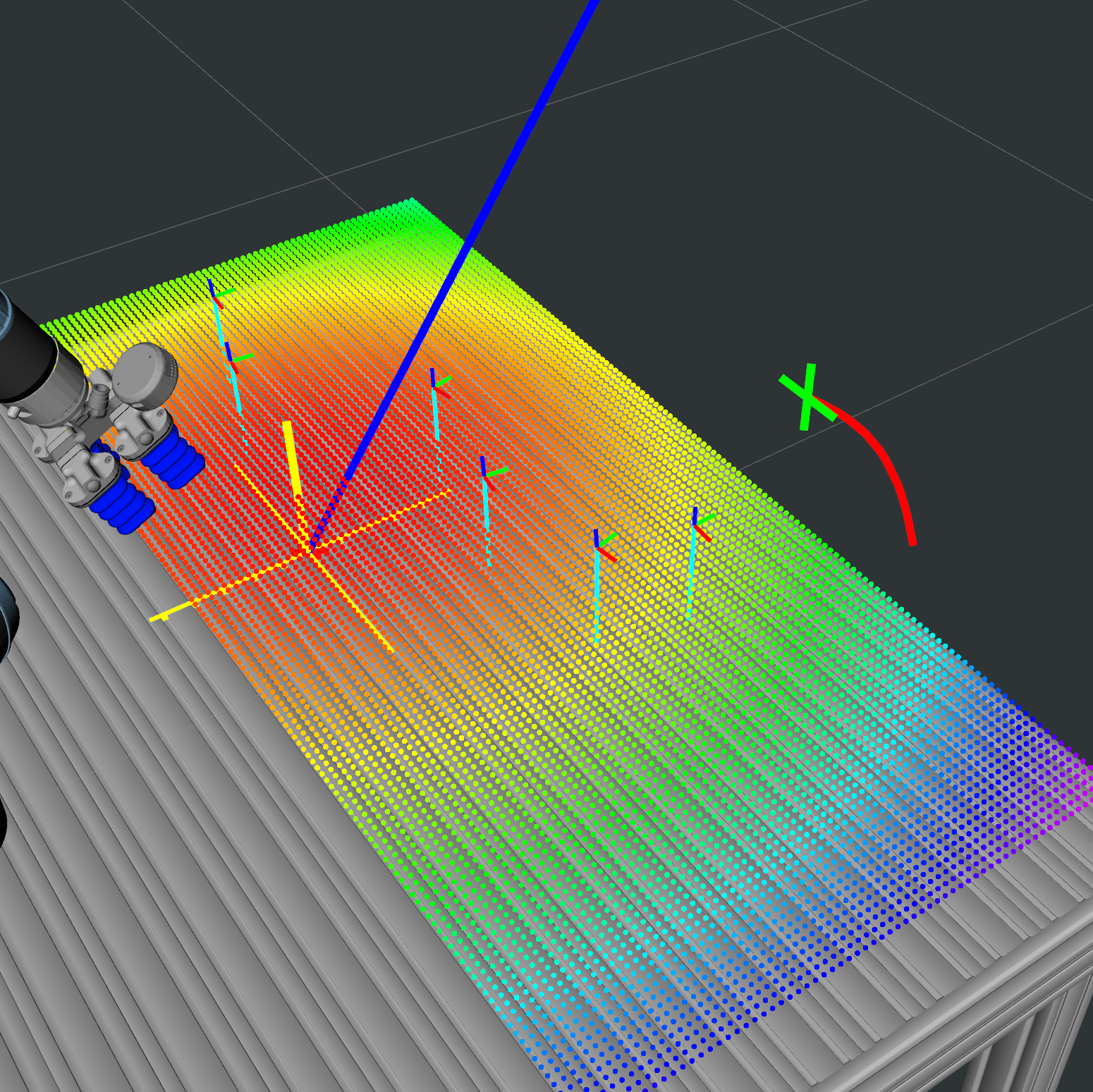}
    \includegraphics[width=0.475\columnwidth]{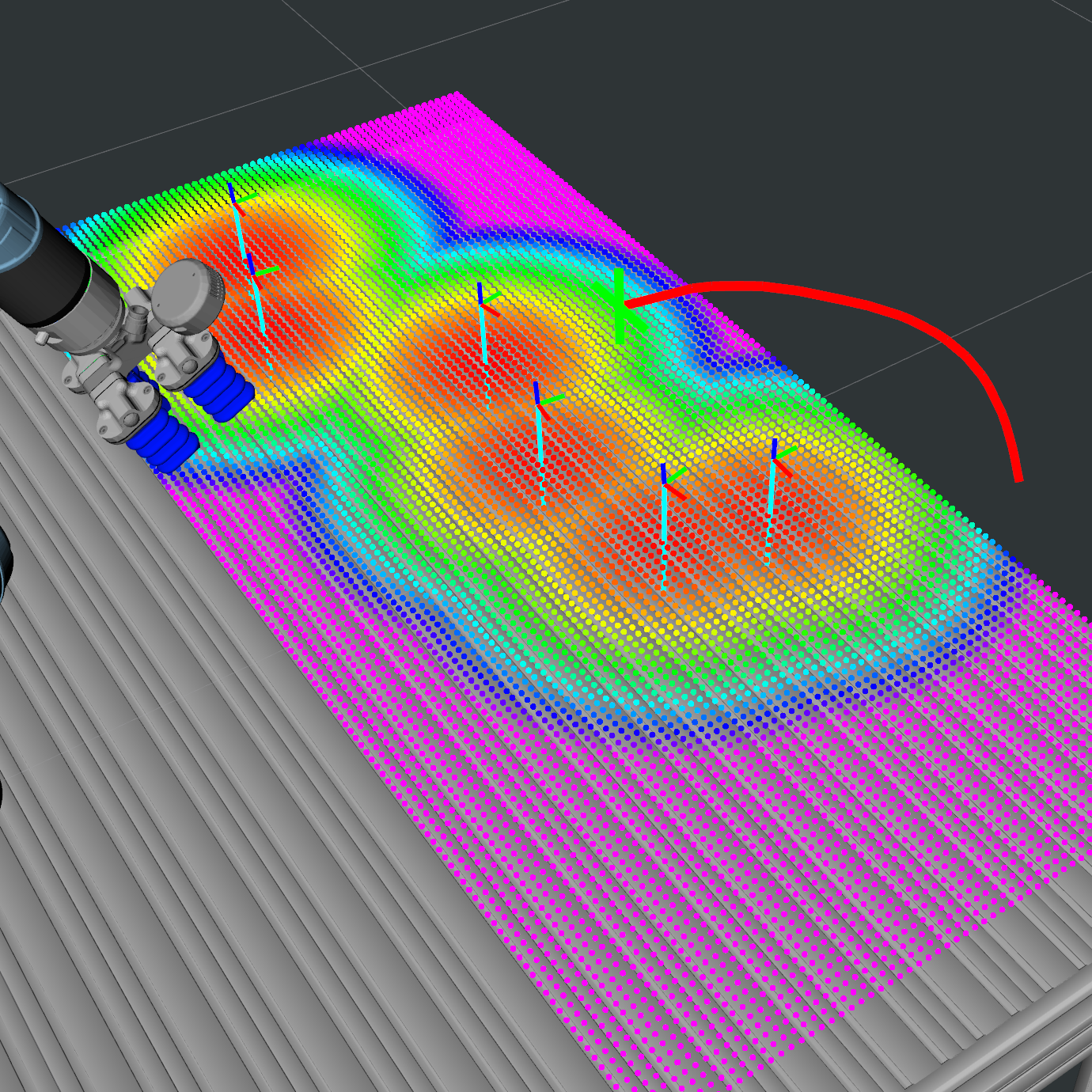}
    \caption{Three different prior distribution examples. Left: Scene layout. Mid-Left: Gaussian centered on the area close to the human. Mid-Right: Gaussian centered on the gaze-table (blue ray) intersection (yellow crosshair). Right: Mixture model with a mode over each detected object.}
    \label{fig:priors_design}
\end{figure*}

One might find it reasonable that, without any other task constraints, humans might favor to reach for locations closer to them. This "common sense" belief from the designer can be encoded into the inference process as a prior probability using a Normal distribution centered on the human. See an example in Figure~\ref{fig:priors_design} Middle-Left where the prior is a $\mathcal{N}(0, I*0.1)$. Other a priori information like, subject handedness or height could be used to modify the distribution location and scale values.

Typically, humans look at the grasping points when performing manipulation \cite{Palinko2016}. Therefore, gaze is a strong prediction cue for human reaching intent. Provided we have gaze tracking capabilities, we can design a prior distribution that favors locations that are close to the looking direction. For our application, we used a Normal distribution centered on the gaze-table intersection. In this case, instead of a fixed covariance model, the distribution scale is proportional to the covariance of the gaze-table intersection points observed over a time window of 0.3s, see Figure~\ref{fig:priors_design} Middle-Right.

Finally, we can include another "common sense" cue. If the task at hand consists of grasping objects, it is likely that the human will reach for an object rather than empty workspace. Since we have a dependable 3D object perception system for the robot, capable of real-time pose estimation with $\pm$5mm precision, object poses can be used as prior information for inferring human's reaching intent. A Gaussian Mixture Model with a mode over each detected object (see Figure~\ref{fig:priors_design} Right) can encode such prior information. The scale parameter of each mixture component can be determined, depending on detection confidence and object size.

All three priors can be combined into a comprehensive prior distribution using a weighted mixture model. Mixture weights increase the flexibility of the priors. Depending on the task, subject or other context conditions different values can be more appropriate. With a few assumptions, a complex informative prior that fuses information from multiple sources is derived. The attached video shows more detail about the prior design. 

\subsection{Real-time inference with neural surrogates and vectorized grid sampling}
With the building blocks described in the previous sections, we can perform intent inference using ABC. However, the approach becomes impractical due to the number of samples that need to be generated and the computation time that each sample implies. For the described generative model, each trajectory generation takes about 150 ms on a high-end desktop CPU, see Section~\ref{ssec:robot} for details about experimental platform. Even considering a crude approximation to the intent posterior and using a low number of samples ($\sim$100), each prediction could take around 15s which defeats the purpose of anticipating the intent and acting preemptively. This problem has been reliably addressed by introducing neural surrogates into the ABC inference pipeline.

\begin{figure}
    \centering
    \includegraphics[width=0.475\columnwidth]{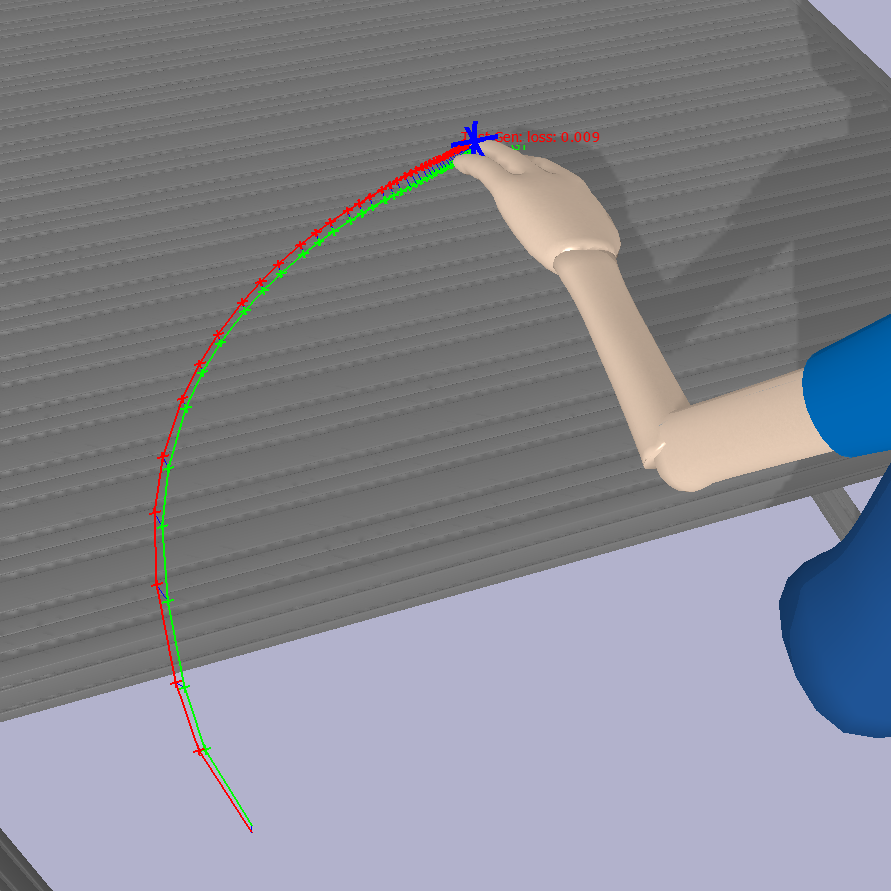}
    \includegraphics[width=0.475\columnwidth]{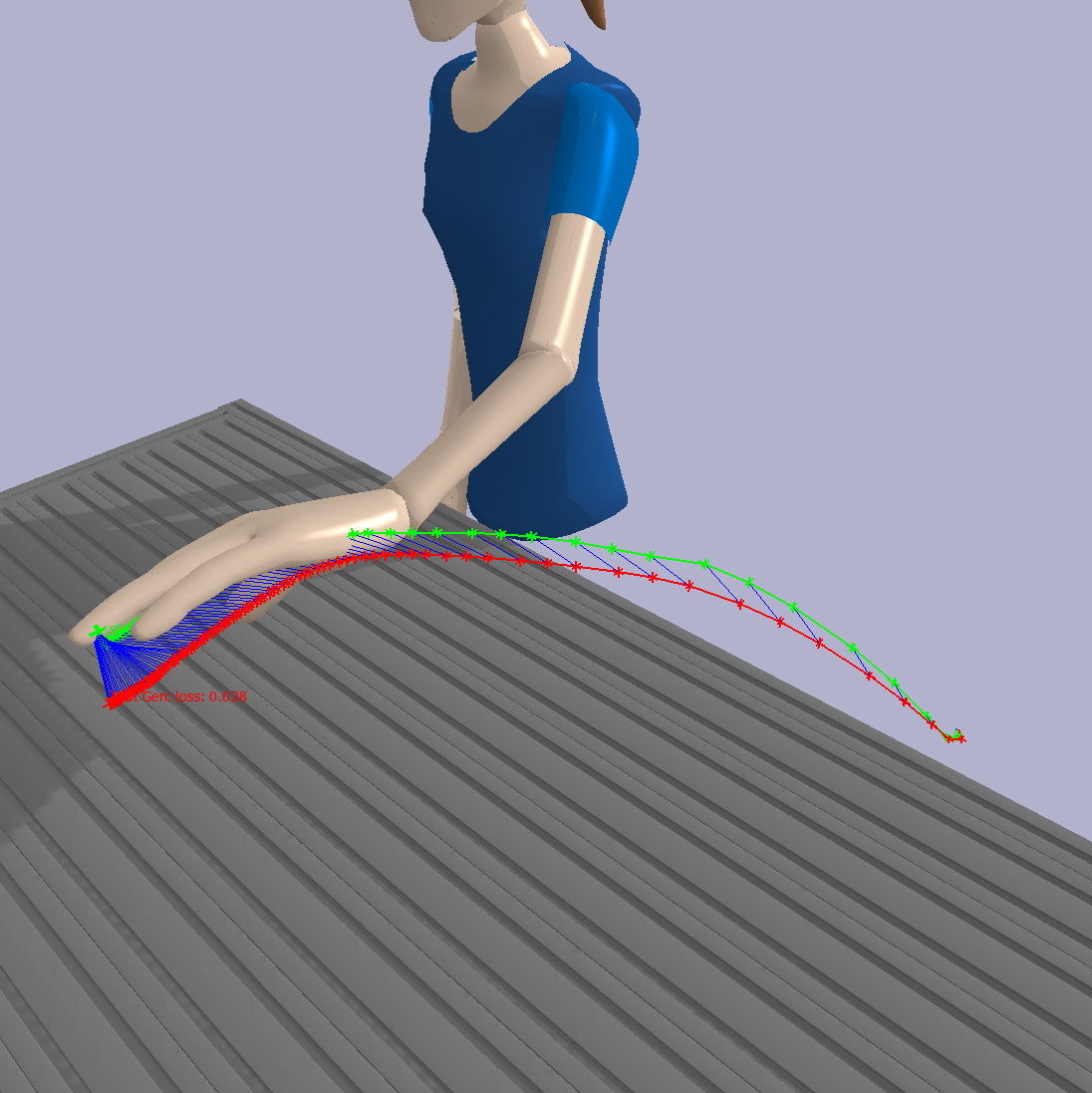}
    \caption{Examples of simulator generated trajectories (green) and their approximations using the trained neural surrogate (red). Point to point differences are shown by blue lines. Shown trajectories have a mean error of 0.9cm (left) and 2.8cm (right).}
    \label{fig:neural_surrogate}
\end{figure}

Thanks to the generative model, it is possible to generate data offline and create a synthetic dataset of human reach intents. The dataset can be used to train a neural network as a function approximator for the generative process, i.e. a neural surrogate. By replacing the generative model by its surrogate, the generative workload becomes a matrix-matrix multiply-accumulate type of compute which can leverage optimizations available in modern CPU/GPUs. The improvement is exacerbated when combining it with stratified sampling approaches, e.g. grid sampling. Vectorized implementations of the ABC components can result in massive sampling and evaluation operations in parallel. In the concrete implementation for intent prediction workload discussed in this paper, the neural surrogate is able to sample and evaluate 9,100 samples in 0.01s instead of 1365s using the original single-threaded, physics simulation generative model. More than a 100,000x improvement. The neural surrogate approximation quality can be seen in Figure~\ref{fig:neural_surrogate}. After the training has converged the neural surrogate mean error is 1.7cm.

The neural surrogate used in this paper consists of a 3-layer fully connected network (FC3x32-FC32x64-FC64x270). It provides good approximations without much complexity, see Figure~\ref{fig:neural_surrogate}. To train it we generated a dataset of 10,000 trajectories using the offline simulator and use a 0.8/0.2 training testing split. The 3 values at the input layer are interpreted as the target reaching position. The 270 output values are interpreted as a trajectory consisting of 90 tridimensional points (3s @ 30Hz). In general, the neural surrogate architecture will depend on the complexity and nature of the generative process.

The neural surrogate lacks the expressive power of a complete physics engine and might not generalize to other environments or kinematics. However, it is always possible to retrain it offline or select a neural surrogate from a library depending on task and context. Intuitively, the neural surrogate learns a simplified version of the physics specialized on generating reaching trajectories for a specific scene. Such simplification results in much faster trajectory generation.

When the generative model is already simple, this approach might still be useful. Replacing a generative model by a neural approximation unlocks matrix-matrix optimizations and single instruction multiple data computations which can reduce computation time by orders of magnitude.

Additionally, if the inference strategy is grid-based the generated trajectories can be cached, further reducing the computation time of subsequent evaluations. In our experiments, the inference space is discretized into 9,100 cells using a regular grid. Intent is evaluated for each cell after each new trajectory point is observed. During the first inference run, a trajectory is generated for each cell and stored resulting in a total evaluation time of 10ms. Subsequent inferences skip trajectory generation and take 1ms.

\section{Human intent inference and task fluency}

\subsection{Task description}
\label{sec:task_desc}

\begin{figure}
    \centering
    \includegraphics[width=0.31\columnwidth, cfbox=reachColor 2pt -1pt]{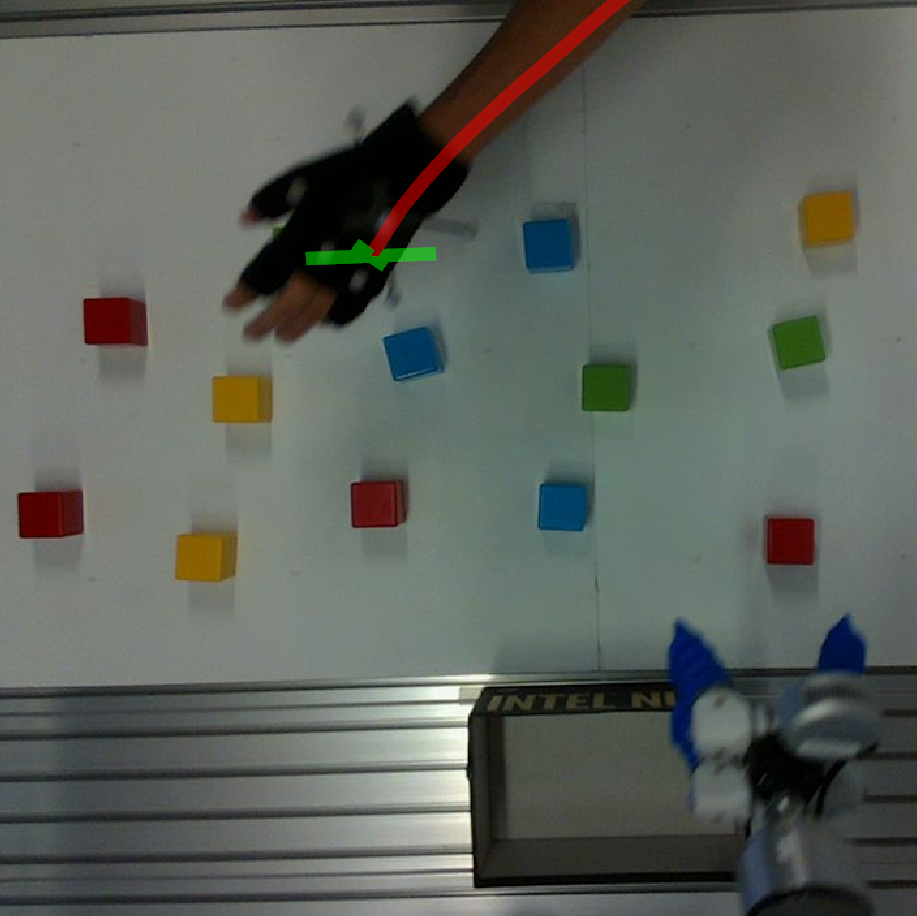}
    \includegraphics[width=0.31\columnwidth, cfbox=transColor 2pt -1pt]{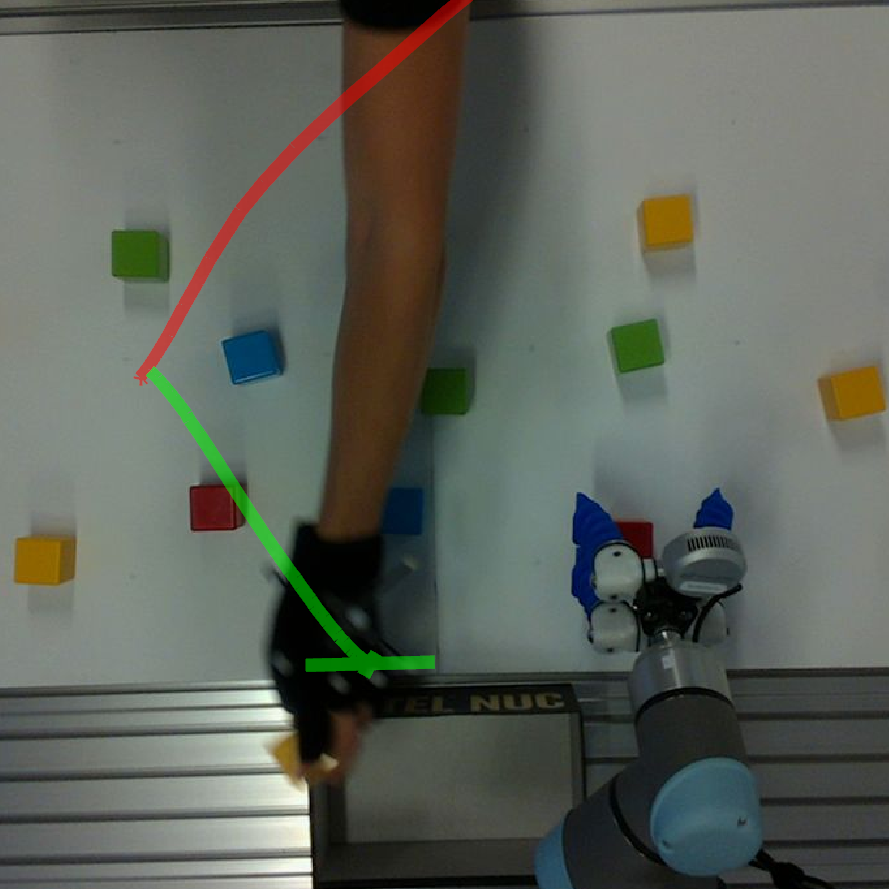}
    \includegraphics[width=0.31\columnwidth, cfbox=retreatColor 2pt -1pt]{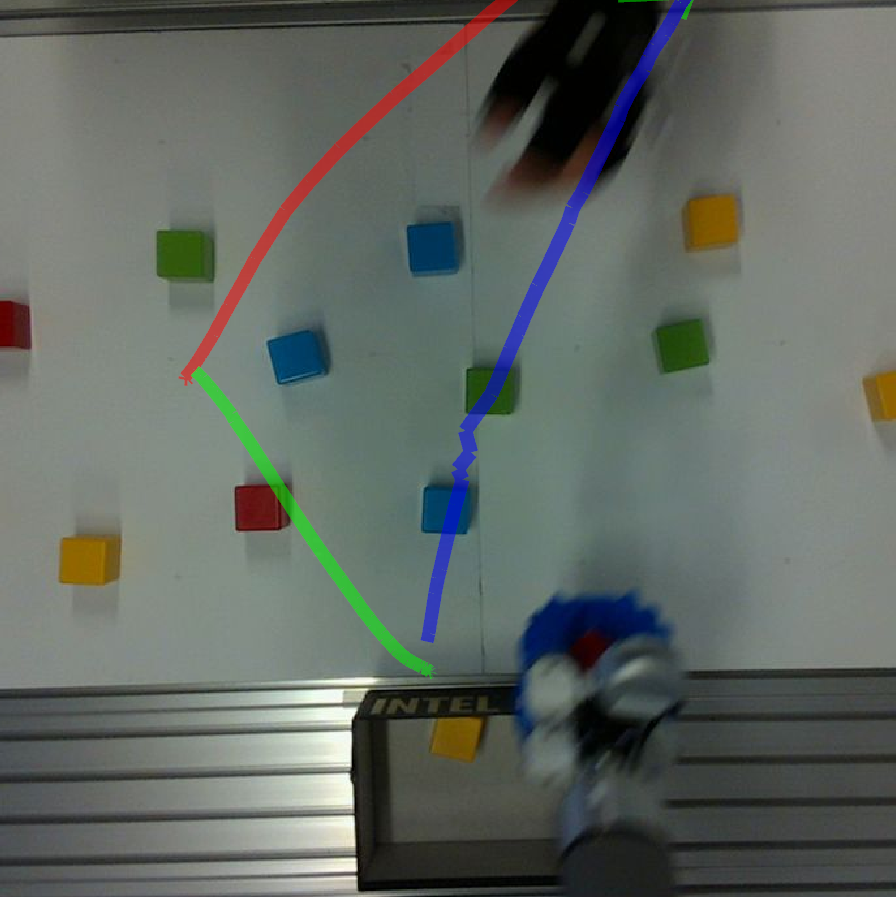}
    \includegraphics[width=\columnwidth]{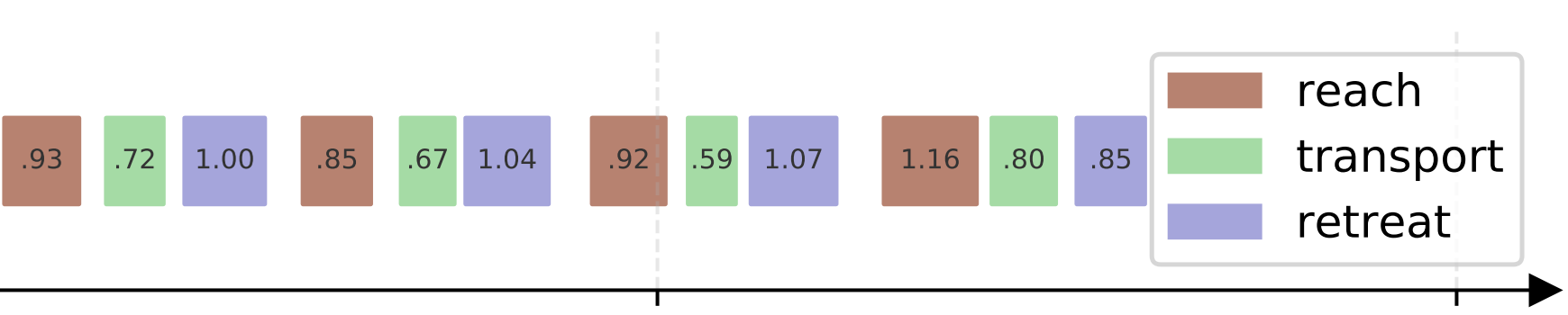}

    \caption{Execution diagram of four pick and place actions each one consisting of a \reach{reach}, \transport{transport}, \retreat{retreat} sequence. Each atomic action is represented by a color-coded block labeled with the time it took to be completed. The thumbnails on the first row show a frame of each color-coded atomic action and its trajectories.}
    \label{fig:task_diagram}
\end{figure}

The task consists of picking and placing 16 objects into a box. The task environment is shown in Fig~\ref{fig:exp_setup}. It consists of a table with a UR5e robot arm on one side and the human on the other. On top of the table, there are 16 colored cubes and an empty box, the objects are at random positions and orientations. The task is considered started when any of the actors starts moving and ends upon the last object being placed into the box. Therefore, there are 16 pick and place sequences required to complete the task. Each pick and place sequence can be split into five atomic actions: \reach{reach}, grasp, \transport{transport}, release and \retreat{retreat}. We visualize task execution as a sequence of atomic actions, see the task diagram in Fig~\ref{fig:task_diagram}. In this paper, grasp and release consist of closing and opening the robot gripper or human hand respectively. To reduce clutter, they are omitted from task diagrams although they contribute to total task time.

\subsection{Improving collaboration with reaching intent prediction}
If multiple agents are performing the task simultaneously, there are multiple potential conflict points. If both agents perform a \transport{transport} action simultaneously, they might run into each other since the box is shared. During the \reach{reach} phase, both agents can try to reach for the same object or for two objects that are close enough to interfere with each other's actions. Typically these conflicts are avoided using a turn-taking approach, where each agent performs one pick and place action at a time. However, strict turn taking increases the idle time of both agents, making them wait for each other for an entire pick and place action.

Intent prediction allows the robot to start its \reach{reach} action once there is enough certainty that the reached object is away from human's intent. If there are multiple candidate objects, it selects the object that is further from any other object that is still likely to be reached. 

\begin{figure*}[ht]
    \centering
    \includegraphics[width=\textwidth]{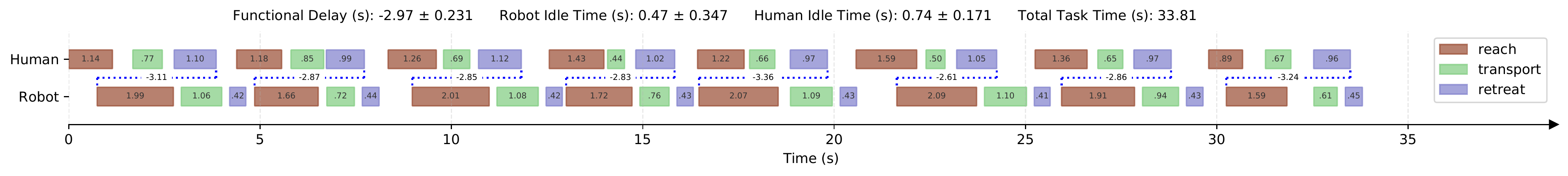}
    \includegraphics[width=\textwidth]{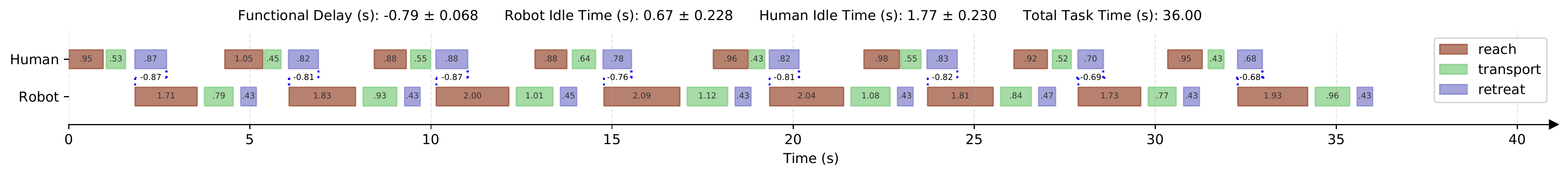}
    \includegraphics[width=\textwidth]{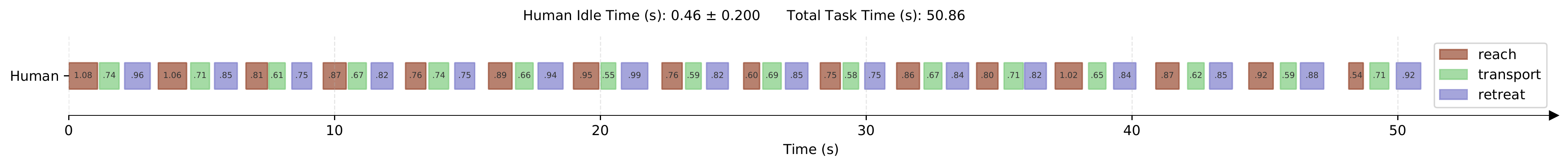}
    \includegraphics[width=\textwidth]{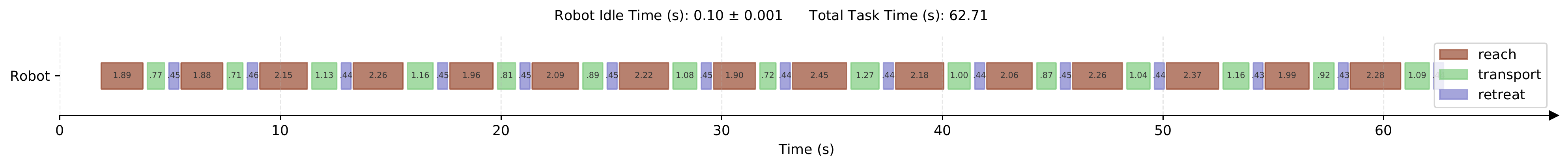}
    \vspace{-8mm}
    \caption{Experimental results depicted as a task phase diagram with fluency metrics. Each row shows a complete task where the 16 objects were picked and placed into the box. First row: Human intent prediction. Second row: Turn taking baseline. Third row: Solo human. Forth row: Solo robot. \emph{This is vector a graphics figure, details are best analyzed zooming in on a digital device.}}
    \label{fig:task_results}
\end{figure*}

This anticipation allows for much concurrent task execution shown in the experimental results in Figure~\ref{fig:task_results}. Once the robot reaches the object it waits to perform its \transport{transport} action until the human starts its \retreat{retreat} action.

\subsection{Measuring collaboration}
In the literature, multiple metrics to evaluate collaboration in human-human and human-robot tasks have been used. So far there is no consensus on which metrics are best. Performance is typically related with the overall time taken to complete a task. However, total task time does not tell the full story and other metrics are needed \cite{Hoffman2019}. Multiple studies point towards task fluency (i.e. perceived collaboration effectiveness) as an important metric for acceptance of collaborators \cite{Butepage2017, Hoffman2019}. Task fluency is a qualitative metric typically measured through experiments and questionnaires. Hoffman identified strong correlations between subjective task fluency and other objective metrics such as functional delay and idle times \cite{Hoffman2019}. In this paper, we focus on measuring total elapsed time (T) for task performance and functional delay (FD), robot idle time (RI) and human idle time (HI) for task fluency. For our specific task and environment the aforementioned metrics are obtained as follows:

\begin{itemize}
    \item \emph{Elapsed time (T)}: Time taken since the human starts \reach{reaching} for the first object until the last object is packed into the box and the corresponding \retreat{retreat} action is finished.
    \item \emph{Functional delay (FD)}: Robot delay to start a \reach{reach} action after the human has completed a \retreat{retreat} action. This is typically due to computation time required to update the scene state and plan an action. Notice that this delay can be negative when the agent is able to anticipate and start moving before the partner agent is done.
    \item \emph{Robot idle time (RI)}: Elapsed time in which the robot is not performing any action.
    \item \emph{Human idle time (HI)}: Elapsed time in which the human is not performing any action.
\end{itemize}

\begin{figure}
    \centering
    \includegraphics[width=0.975\columnwidth]{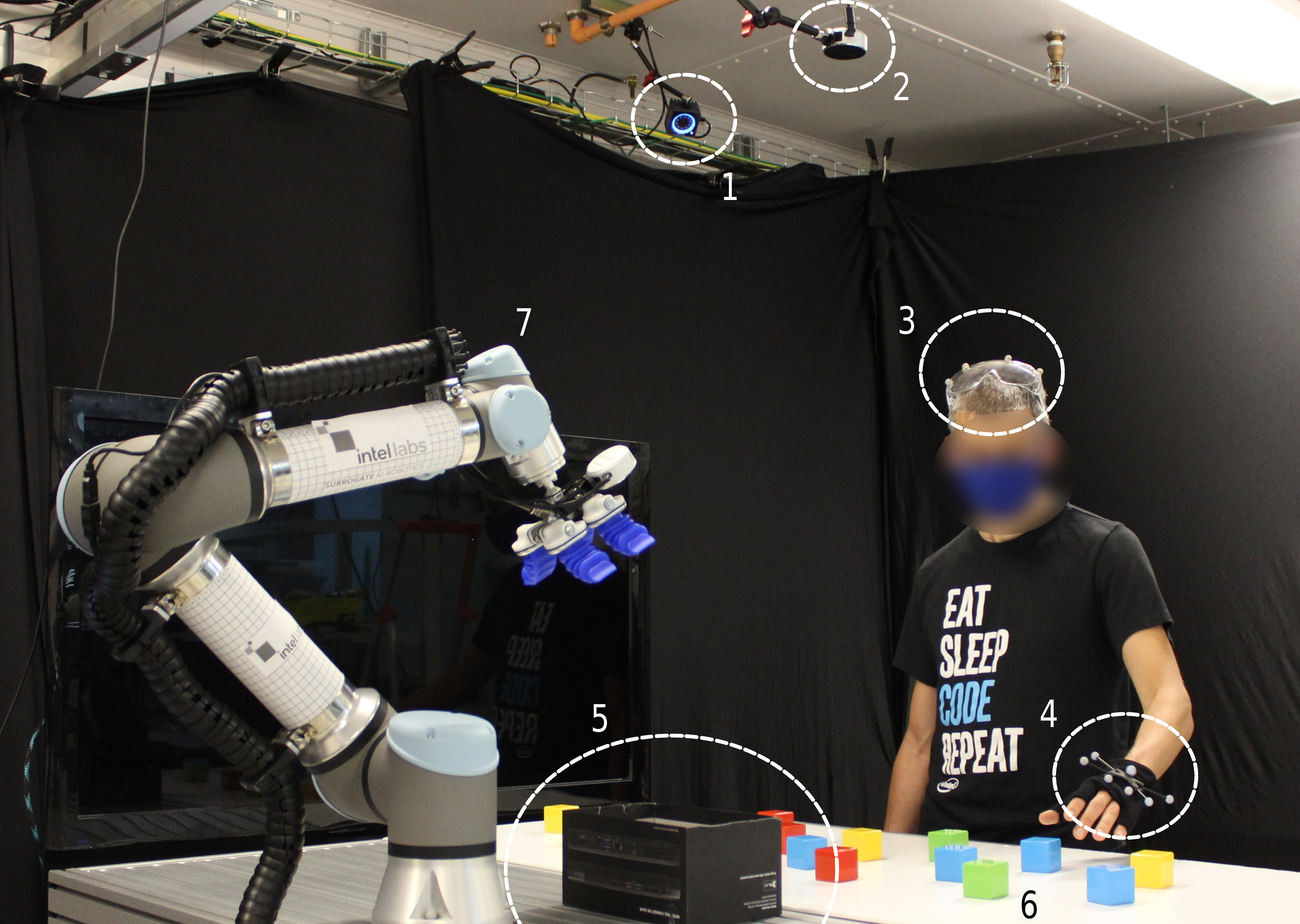}
    \caption{Experimental setup for human-robot collaboration. The UR5 cobot with an mGrip soft gripper on the left (7) and the human on the right collaborate to pick all objects and place them into the box (5). Objects are laid out randomly on the workspace (6). A motion capture system (1) tracks the human looking direction and hand motion using markers (3, 4) and a Intel Realsense L515 camera (2) is used to detect the objects pose.}
    \label{fig:exp_setup}
\end{figure}

\begin{table}
\centering
\caption{Task performance and fluency metrics results}
\label{tab:results}
\begin{tabular}{lllll}
 &
  \begin{tabular}[c]{@{}l@{}}Solo \\ Human\end{tabular} &
  \begin{tabular}[c]{@{}l@{}}Solo \\ Robot\end{tabular} &
  \begin{tabular}[c]{@{}l@{}}Collab \\ Baseline\end{tabular} &
  \begin{tabular}[c]{@{}l@{}}Intent \\ Prediction\end{tabular} \\ \cline{1-5} 
Time & 57.05 $\pm$4.49      & 64.55$ \pm$1.33    & 40.69 $\pm$6.68  & 36.49 $\pm$2.78\\
FD   & n/a        & n/a        & -0.85 $\pm$0.07           & -2.91 $\pm$0.26 \\
RI   & n/a        & n/a        & 1.09 $\pm$0.62            & 0.62 $\pm$0.22 \\
HI   & n/a        & n/a        & 2.02 $\pm$0.74            & 0.97 $\pm$0.28 \end{tabular}
\end{table}

\section{Experiments}
We performed 10 task executions for each evaluated method. Each task consisting on pick and placing 16 objects into the box results in a total of 160 pick and place sequences per method evaluated. Before each experiment, objects were laid out on the table without any specific order or position, the only constraint was to keep them in the workspace of both the robot and the human and separated about 5 cm from each other. Velocity profiles of the human hand obtained by the motion capture system are used to segment the different phases of the pick and place sequence performed by the human (See Figure~\ref{fig:task_diagram}). The human was not given any specific instructions besides the task description.





\subsection{Baselines}
To evaluate the impact of intent prediction to anticipate the robot's \reach{reach} action we propose three different baselines. The first two baselines are non-collaborative task executions where the human and the robot perform the task solo. This provides a clear and objective insight of the performance that a single agent can achieve by itself. 

Each agent is on a different table end, therefore \retreat{retreat} actions are unlikely to cause conflicts. The collaborative baseline we propose exploits this fact and starts the robot's \reach{reach} action when the human starts its \retreat{retreat} action. This baseline does not need to perform any intent prediction, just keep track of what action the human is performing. In the results section, it can be seen how the collaborative baseline substantially improves metrics compared to the solo baselines.

\subsection{Robot system}
\label{ssec:robot}
The robotic platform consists of a UR5e arm equipped with a Soft Robotics mGrip soft gripper, see Fig~\ref{fig:exp_setup}. The robot is steered by a joint velocity controller. The end-effector is moved to the desired reach and transport positions by means of differential kinematics and a PD controller on the Cartesian space. The perception system consists of an optitrack motion capture system and a Realsense L515 depth camera. The optitrack system, consisting of four cameras, is used to obtain the human hand pose and the human looking direction. The L515 camera is used for object pose estimation and obstacle detection. See Fig~\ref{fig:exp_setup}. All evaluated methods and experiments are computed on an Intel(R) NUC-8i7HVK with an Intel(R) Core(TM) i7-8809G CPU @ 3.10GHz and 32GB DDR4 RAM.

\section{Results and discussion}
Table~\ref{tab:results} provides the summary statistics for the evaluated methods. It can be seen how leveraging intent prediction improves the baselines on all metrics. This makes the task finish earlier, reduces idle times and improves task fluency by greatly reducing functional delay.

Figure~\ref{fig:task_results} shows a task diagram from the best run of each of the methods. By looking at the overall structure of each diagram, one can notice that the blocks are much more compact when prediction is used. This is indeed reflected on the metrics by reducing the idle times and substantially decreasing the functional delay. The improvement in task fluency can be better noticed in the video attachment. In the video attachment, task diagrams for all the experiments are shown.

\section{Conclusion}
In this paper we have presented an approach to perform human reaching intent prediction. We show how to formulate the problem into the Approximate Bayesian Computation framework, a principled likelihood-free technique that allows for great flexibility on the model design and eases the inclusion of prior knowledge and sensor data. We have proposed the necessary optimizations to allow ABC to run at interactive frame rates. Next, the predicted intent is leveraged to improve task fluency on a collaborative pick and place task. The experimental evaluation supports that human intent is key to improve task fluency. Predicting where a human might reach next can be leveraged for other collaborative tasks like packing, assembly, etc.



\IEEEtriggeratref{9}
\bibliographystyle{IEEEtran}
\bibliography{biblio.bib}

\end{document}